\pdfoutput=1

\documentclass[11pt]{article}

\usepackage[]{acl}

\usepackage{times}
\usepackage{latexsym}

\usepackage[T1]{fontenc}

\usepackage[utf8]{inputenc}

\usepackage{microtype}
\usepackage{graphicx}
\usepackage{subfigure}
\usepackage{amsmath}
\usepackage{amssymb}
\usepackage{svg}
\usepackage{multirow}
\usepackage{enumitem}
\usepackage{booktabs}
\usepackage{algorithm}
\usepackage{algorithmic}

\title{Improving Robustness of Language Models from a Geometry-aware Perspective}

\author{Bin Zhu${^{1}}$ , Zhaoquan Gu${^{1,2}}$\thanks{\ \ Corresponding author} , Le Wang${^{1,2}}$, Jinyin Chen${^{3}}$, Qi Xuan${^{3}}$ \\
  \small $^1$ Cyberspace Institute of Advanced Technology (CIAT), Guangzhou University, Guangzhou 510006, China\\
  \small $^2$ Institute of Cyberspace Platform, Peng Cheng Laboratory, Shenzhen 999077, China\\
  \small $^3$ Institute of Cyberspace Security, Zhejiang University of Technology, Hangzhou 310023, China\\
  {\small \tt zhubin@e.gzhu.edu.cn, \{zqgu,wangle\}@gzhu.edu.cn} \\ {\small \tt \{chenjinyin,xuanqi\}@zjut.edu.cn} \\
}

\begin{document}
\maketitle
\begin{abstract}
Recent studies have found that removing the norm-bounded projection and increasing search steps in adversarial training can significantly improve robustness.
However, we observe that a too large number of search steps can hurt accuracy.
We aim to obtain strong robustness efficiently using fewer steps. 
Through a toy experiment, we find that perturbing the clean data to the decision boundary but not crossing it does not degrade the test accuracy.
Inspired by this, we propose friendly adversarial data augmentation (FADA) to generate friendly adversarial data.
On top of FADA, we propose geometry-aware adversarial training (GAT) to perform adversarial training on friendly adversarial data so that we can save a large number of search steps. 
Comprehensive experiments across two widely used datasets and three pre-trained language models demonstrate that GAT can obtain stronger robustness via fewer steps.
In addition, we provide extensive empirical results and in-depth analyses on robustness to facilitate future studies.
\end{abstract}

\section{Introduction}
\label{sec:intro}
Deep neural networks (DNNs) have achieved great success on many natural language processing (NLP) tasks \cite{kim-2014-convolutional,vaswani-etal-2017-attention,devlin-etal-2019-bert}. 
However, recent studies \cite{szegedy-etal-2013-intriguing,goodfellow-etal-2014-explaining} have shown that DNNs are vulnerable to crafted adversarial examples . 
For instance, an attacker can mislead an online sentiment analysis system by making minor changes to the input sentences \cite{papernot-etal-2016-crafting,liang-etal-2017-deep}.
It has raised concerns among researchers about the security of DNN-based NLP systems. 
As a result, a growing number of studies are focusing on enhancing robustness to defend against textual adversarial attacks \cite{jia-etal-2019-certified,ye-etal-2020-safer,jones-etal-2020-robust,zhu-etal-2020-freelb}. 

Existing adversarial defense methods fall into two categories: empirical and certified defenses.
Empirical defenses include gradient-based adversarial training (AT) and discrete adversarial data augmentation (ADA). 
Certified defenses provide a provable guaranteed robustness boundary for NLP models. This work focuses on empirical defenses.

There was a common belief that gradient-based AT methods in NLP was ineffective compared with ADA in defending against textual adversarial attacks \cite{li-and-qiu-2021-token,si-etal-2021-better}. 
\citet{li-etal-2021-searching} find that removing the norm-bounded projection and increasing the number of search steps in adversarial training can significantly improve robustness.
Nonetheless, we observe that increasing the number of search steps further does not significantly improve robustness but hurts accuracy.

\begin{figure}
    \centering
    \includegraphics[width=\columnwidth]{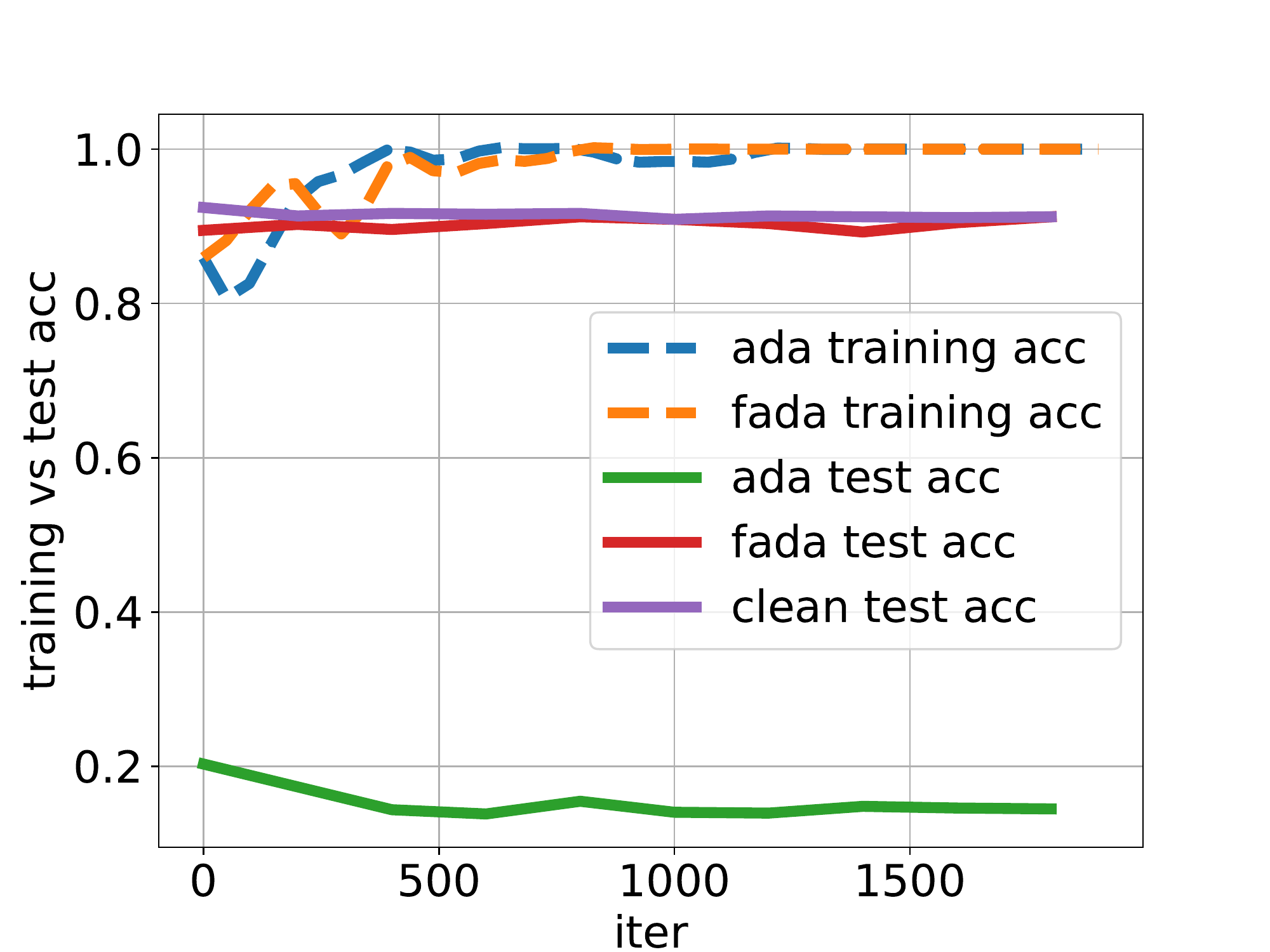}
    \caption{The clean accuracy achieved with ADA, FADA, and the original training set. During training, both ADA and FADA have close to 100\% accuracy. However, ADA only achieves $\sim$15\% accuracy during testing while FADA maintains the same test accuracy with the original training set. This indicates that training data which crosses the decision boundary hurts the accuracy significantly.}
    \label{fig:training_test_acc}
\end{figure}

\begin{figure}
    \centering
    \includegraphics[width=\columnwidth]{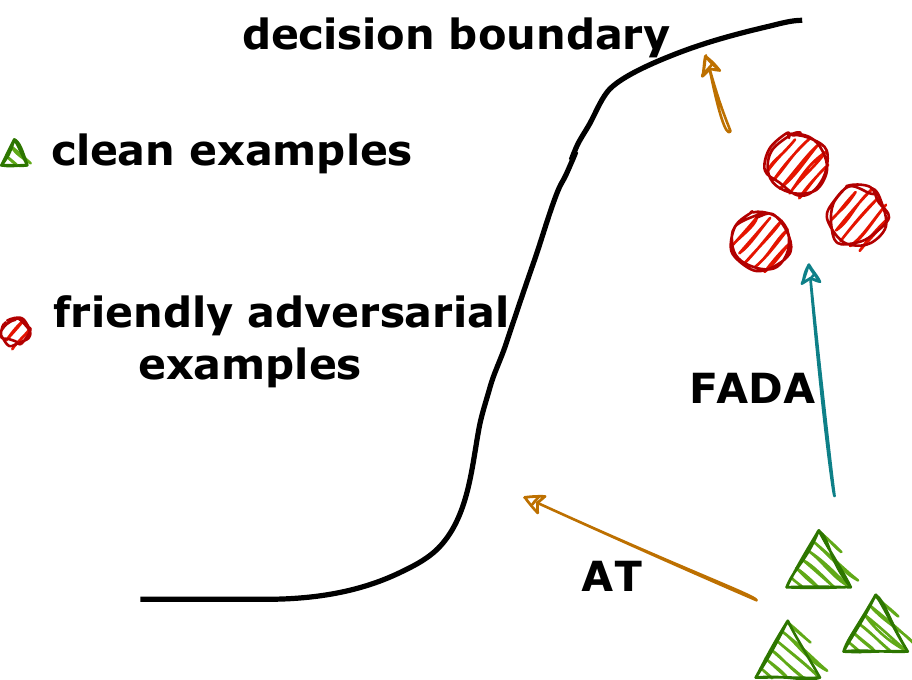}
    \caption{Illustration of GAT. Our GAT can save many search steps since friendly adversarial examples are located near the decision boundary.}
    \label{fig:illustrate}
\end{figure}

We give a possible explanation from a geometry-aware perspective. 
Removing the norm-bounded projection enlarge the search space.
Appropriately increasing the number of search steps brings the adversarial data closer to the decision boundary.
In this case, the model learns a robust decision boundary. 
Further increasing the number of search steps can make the adversarial data cross the decision boundary too far, hindering the training of natural data and hurting natural accuracy.

To verify our hypothesis, we train a base model using adversarial data, which are generated by adversarial word substitution (AWS) on the SST-2 \cite{socher-etal-2013-recursive} dataset. 
We report its training accuracy (``ada training acc'') on adversarial data and test accuracy (``ada test acc'') on the clean test set in Figure~\ref{fig:training_test_acc}. 
Although achieving nearly 100\% training accuracy, its test accuracy is only about 15\%, which implicates the adversarial data make the test performance degraded. 
Then we train another base model, whose training data is more ``friendly''. We just recover their last modified words to return to the correct class, namely friendly adversarial data augmentation (FADA).
It means that only one word is different in each sentence.
Surprisingly, it achieves a high test accuracy of $\sim$93\%.

This preliminary inspired us to address two existing problems:
\begin{itemize}
    \item \textbf{The number of search steps is always large, which is computationally inefficient}.
    \item \textbf{A too large number of steps leads to degraded test performance}.
\end{itemize}

Geometrically speaking, the friendly adversarial data are close to the ideal decision boundary. 
We can address the above two issues in one fell swoop if we perform gradient-based adversarial training on these friendly adversarial data. 
It is like we start one step before the end, allowing us to obtain strong robustness through a tiny number of search steps.
We name it geometry-aware adversarial training (GAT). 
Figure~\ref{fig:illustrate} illustrates our proposed GAT.


In addition, the friendly adversarial data only need to be generated once per dataset. It can be reused, so it is computationally efficient. 
It can also be updated for every iteration or epoch but computationally expensive.

Our contributions are summarized as follows:
\begin{itemize}
    \item [1)] We propose FADA to generate friendly adversarial data which are close to the decision boundary (but not crossing it).
    \item [2)] We propose GAT, a geometry-aware adversarial training method that adds FADA to the training set and performs gradient-based adversarial training. 
    \item [3)] GAT is computationally efficient, and it outperforms state-of-the-art baselines even if using the simplest FGM. We further provide extensive ablation studies and in-depth analyses on GAT, contributing to a better understanding of robustness.
\end{itemize}


\section{Related Work}
\label{sec:r_w}
\subsection{Standard Adversarial Training}
Let $f_\theta(x)$ be our neural network, $\mathcal{L}(f_\theta(x),y)$ be the loss function (e.g., cross entropy), where $x\in X$ is the input data and $y\in Y$ is the true label.
The learning objective of standard adversarial training is
\begin{equation}
    \mathop{\min}_{\theta} \mathbb{E}_{(X,Y)\sim D}\left[\mathop{\max}_{\|\delta\|\leq\epsilon} \mathcal{L}(f_\theta(X+\delta),y)\right],
    \label{eq:learning_obj}
\end{equation}
where $D$ is the data distribution, $\delta$ is the minor perturbation, $\epsilon$ is the allowed perturbation size.
To optimize the intractable min-max problem, we search for the optimal $\delta$ to maximize the inner loss and then minimize the outer loss w.r.t the parameters $\theta$, step by step.

The gradient $g$ of the inner loss w.r.t the input $x$ is used to find the optimal perturbation $\delta$.
\citet{goodfellow-etal-2014-explaining} proposed fast gradient sign method (FGSM) to obtain $\delta$ by one step:
\begin{equation}
    \delta = \epsilon \cdot sgn(g),
    \label{eq:fgsm}
\end{equation}
where $sgn(\cdot)$ is the signum function.
\citet{Madry-etal-2018-towards} proposed projected gradient descent (PGD) to solve the inner maximization as follows:
\begin{equation}
    \delta^{\left(t+1\right)}=\Pi\ \alpha\cdot g^{(t)}/\|g^{(t)}\|, \forall t \geq 0,
    \label{eq:pgd}
\end{equation}
where $\alpha > 0$ is the step size (i.e., adversarial learning rate), $\Pi$ is the projection function that projects the perturbation onto the $\epsilon$-norm ball. Conventionally PGD stops after a predefined number of search steps $K$, namely PGD-$K$. In addition, TRADES \cite{zhang-etal-2019-theoretically}, MART \cite{Wang-etal-2020-improving-adversarial} and FAT \cite{zhang-etal-2020-attacks} are also effective adversarial training methods for boosting model robustness. 

Regarding FAT, the authors propose to stop adversarial training in a predefined number of steps after crossing the decision boundary, which is a little different from our definition of ``friendly''.

\subsection{Adversarial Training in NLP}
Gradient-based adversarial training has significantly improved model robustness in vision, while researchers find it helps generalization in NLP.
\citet{miyato-etal-2017-adversarial} find that adversarial and virtual adversarial training have good regularization performance.
\citet{sato-etal-2018-interpretable} propose an interpretable adversarial training method that generates reasonable adversarial texts in the embedding space and enhance models' performance. 
\citet{zhu-etal-2020-freelb} develop FreeLB to improve natural language understanding. 

There is also a lot of work focused on robustness. 
\citet{wang-etal-2021-infobert} improve model robustness from an information theoretic perspective. 
\citet{dong-etal-2021-towards} use a convex hull to capture and defense against adversarial word substitutions. 
\citet{zhou-etal-2021-defense} train robust models by augmenting training data using Dirichlet Neighborhood Ensemble (DNE).

Besides, adversarial data augmentation is another effective approach to improve robustness \cite{ebrahimi-etal-2018-hotflip,li-etal-2018-textbugger,ren-etal-2019-generating,jin-etal-2019-is-bert,zang-etal-2020-word,li-etal-2020-bert-attack,garg-and-ramakrishnan-2020-bae,si-etal-2021-better}. However, it only works when the augmentation happens to be generated by the same attacking method and often hurts accuracy. 

It is worth noting that recent empirical results have shown that previous gradient-based adversarial training methods have little effect on defending against textual adversarial attacks \cite{li-etal-2021-searching,si-etal-2021-better}. The authors benchmark existing defense methods and conclude that gradient-based AT can achieve the strongest robustness by removing the norm bounded projection and increasing the search steps.

\section{Methodology}
\label{sec:method}
\subsection{Friendly Adversarial Data Augmentation}
\label{subsec:fada}

\begin{algorithm}[t]
    \caption{Friendly Adversarial Data Augmentation (FADA)}
    \label{alg:fada}
    \begin{algorithmic}[1]
    \REQUIRE{The original text $x$, ground truth label $y_{true}$,
	    base model $f_{\theta}$, adversarial word substitution function $AWS(\cdot)$}
    \ENSURE{The friendly adversarial example $x_{f}$}
    \STATE Initialization: 
    \STATE $x_f\leftarrow x$
    \STATE the last modified word $w^*$ $\leftarrow$ None
    \STATE the last modified index $i^*$ $\leftarrow$ 0
    \STATE $x_{adv}, w^*, i^*=AWS(x,y_{true},f_{\theta})$
    \IF{$w^*=$ None}
    \RETURN $x_f$

    \ENDIF
    \STATE Replace $w_{i^*}$ in $x_{adv}$ with $w^*$ 
    \STATE $x_f\leftarrow x_{adv}$
    \RETURN \textbf{$x_{f}$}
    
    \end{algorithmic}
\end{algorithm}

For a sentence $x\in X$ with a length of $n$, it can be denoted as $x=w_1w_2...w_i...w_{n-1}w_n$, where $w_i$ is the $i$-th word in $x$. 
Its adversarial counterpart $x_{adv}$ can be denoted as $w_1'w_2'...w_i'...w_{n-1}'w_n'$. 
In this work, $x_{adv}$ is generated by adversarial word substitution, so $x_{adv}$ has the same length with $x$.
Conventional adversarial data augmentation generates adversarial data fooling the victim model and mixes them with the original training set.
As we claim in section~\ref{sec:intro}, these adversarial data can hurt test performance. 
An interesting and critical question is \textbf{when it becomes detrimental to test accuracy}.

\begin{algorithm}
    \small 
    \caption{Ideal Geometry-aware Adversarial Training (GAT)}
    \label{alg:gat}
    \begin{algorithmic}[1]
    \REQUIRE{Our base network $f_{\theta}$, cross entropy loss $\mathcal{L}_{CE}$, training set $D=\{x_i,y_i\}_{i=1}^{n}$, number of epochs $T$, batch size $m$, number of batches $M$}
    \ENSURE{robust network $f_{\theta}$}
    \FOR{epoch = 1 $\TO~T$}
    \FOR{batch = 1 $\TO~M$}
    \STATE Sample a mini-batch $b=\{(x_i,y_i)\}_{i=1}^{m}$
    \FORALL{$x_i$ in $b$}
    \STATE Generate friendly adversarial example $x_i^f$ via Algorithm~\ref{alg:fada}
    \STATE Apply an adversarial training method (e.g., FreeLB++) on both $x_i$ and $x_i^f$ to obtain their adversarial counterpart $\widetilde{x}_i$ and $\widetilde{x}_i^f$
    \ENDFOR
    \STATE Update $f_{\theta}$ via $\nabla_{x}\mathcal{L}_{CE}(f_{\theta}(\widetilde{x}_i),y_i)$ and $\nabla_{x}\mathcal{L}_{CE}(f_{\theta}(\widetilde{x}_i^f),y_i)$
    \ENDFOR
    \ENDFOR
    
    \end{algorithmic}
\end{algorithm}

One straightforward idea is to recover all the $x_{adv}$ to $x$ word by word and evaluate their impact on test accuracy.
We train models only with these adversarial data and test models with the original test set.
We are excited that the test accuracy immediately returns to the normal level when we recover the last modified word.
We denote these data with only one word recovered as $x_f$.
Geometrically, the only difference between $x_{adv}$ and $x_f$ is whether they have crossed the decision boundary.

To conclude, when the adversarial data cross the decision boundary, they become incredibly harmful to the test performance. We name all the $x_f$ as friendly adversarial examples (FAEs) because they improve model robustness without hurting accuracy. Similarly, we name the generation of FAEs as friendly adversarial data augmentation (FADA). We show our proposed FADA in Algorithm~\ref{alg:fada}.

\subsection{Geometry-aware Adversarial Training}
\label{subsec:gat}

\subsubsection{Seeking for the optimal $\delta$}
\label{subsubsec:seek_delta}
Recall the inner maximization issue of the learning objective in Eq.~\eqref{eq:learning_obj}. Take PGD-$K$ as an instance. It divides the search for the optimal perturbation $\delta$ into $K$ search steps, and each step requires a backpropagation (BP), which is computationally expensive. 

We notice that random initialization of $\delta^{0}$ is widely used in adversarial training, where $\delta^{0}$ is always confined to a $\epsilon$-ball centered at $x$.
However, we initialize the clean data via discrete adversarial word substitution in NLP. 
It is similar to data augmentation (DA), with the difference that we perturb clean data in the direction towards the decision boundary, whereas the direction of data augmentation is random.

By doing so, we decompose the $\delta$ into two parts, which can be obtained by word substitution and gradient-based adversarial training, respectively.
We denote them as $\delta_{l}$ and $\delta_{s}$. 
Therefore, the inner maximization can be reformulated as
\begin{equation}
    \mathop{\max}_{\|\delta_{l}+\delta_{s}\|\leq\epsilon} \mathcal{L}(f_\theta(X+\delta_{l}+\delta_{s}),y).
    \label{eq:inner_max_re}
\end{equation}

We aim to find the maximum $\delta_{l}$ that helps improve robustness without hurting accuracy.
As we claim in Section~\ref{subsec:fada}, FADA generates friendly adversarial data which are close to the decision boundary. 
Furthermore, the model trained with these friendly adversarial data keeps the same test accuracy as the original training set (Figure~\ref{fig:training_test_acc}).
Therefore we find the maximum $\delta_{l}$ which is harmless to the test accuracy through FADA.

Denote $X_{f}$ as the friendly adversarial data generated by FADA, Eq.~\eqref{eq:inner_max_re} can be reformulated as
\begin{equation}
    \mathop{\max}_{\|\delta_{s}\|\leq\epsilon} \mathcal{L}(f_\theta(X_{f}+\delta_{s}),y).
    \label{eq:inner_max_re_re}
\end{equation}
The tiny $\delta_{s}$ can be obtained by some gradient-based adversarial training methods (e.g., FreeLB++ \cite{li-etal-2021-searching}) in few search steps. As a result, a large number of search steps are saved to accelerate adversarial training. 
We show our proposed geometry-aware adversarial training in Algorithm~\ref{alg:gat}.

\subsubsection{Final Learning Objective}
\label{subsubsec:lo}

It is computationally expensive to update friendly adversarial data for every mini-batch.
In practice, we generate static augmentation ($X_f$,Y) for the training dataset (X,Y) and find it works well with GAT.
The static augmentation ($X_f$,Y) is reusable.
Therefore, GAT is computationally efficient.

Through such a tradeoff, our final objective function can be formulated as 
\begin{equation}
    \begin{aligned}
    \mathcal{L}= & \mathcal{L}_{CE}(X,Y,\theta)\\& + \mathcal{L}_{CE}(\widetilde{X},Y,\theta)+\mathcal{L}_{CE}(\widetilde{X}_f,Y,\theta),
    \end{aligned}
    \label{eq:final_obj}
\end{equation}
where $\mathcal{L}_{CE}$ is the cross entropy loss, $\widetilde{X}$ and $\widetilde{X}_f$ are generated from $X$ and $X_f$ using gradient-based adversarial training methods, respectively.

\section{Experiments}
\label{sec:exp}

\subsection{Datasets}
\label{subsec:datasets}

We conduct experiments on the SST-2 \cite{socher-etal-2013-recursive} and IMDb \cite{maas-etal-2011-learning} datasets which are widely used for textual adversarial learning. Statistical details are shown in Table~\ref{tab:stat}. We use the GLUE \cite{wang-etal-2019-glue} version of the SST-2 dataset whose test labels are unavailable. So we report its accuracy on the develop set in our experiments.

\begin{table}[htbp]
\setlength {\belowcaptionskip} {-0.5cm}
  \centering
  
    \begin{tabular}{lccc}
    \toprule\toprule
    \bf Dataset & \bf \# train & \bf \# dev / test & \bf avg. length \\
    SST-2 & 67349 & 872   & 17 \\
    IMDb  & 25000 & 25000 & 201 \\
    \bottomrule\bottomrule
    \end{tabular}%
    \caption{Summary of the two datasets.}
  \label{tab:stat}%
\end{table}%

\begin{table*}[htbp]\small 
\setlength{\tabcolsep}{1.2mm}
  \centering

    \begin{tabular}{lcccccccccc}
    \toprule\toprule
    \multicolumn{1}{c}{\multirow{2}{*}{\textbf{SST-2}}} & \multirow{2}{*}{\textbf{Clean \%}} & \multicolumn{3}{c}{\textbf{TextFooler}} & \multicolumn{3}{c}{\textbf{TextBugger}} & \multicolumn{3}{c}{\textbf{BAE}} \\
\cmidrule{3-11}          &       & \textbf{RA \%} & \textbf{ASR \%} & \textbf{\# Query} & \textbf{RA \%} & \textbf{ASR \%} & \textbf{\# Query} & \textbf{RA \%} & \textbf{ASR \%} & \textbf{\# Query} \\
    \midrule
    BERT$_{base}$ & 92.4  & 32.8  & 64.1  & 72.8  & 38.5  & 57.8  & 44.3  & 39.8  & 56.5  & 64.0  \\
    \midrule
    ADA   & 92.2  & 46.7  & 48.7  & 79.4  & 42.0  & 53.9  & 47.0  & 41.2  & 54.8  & 64.0  \\
    ASCC  & 87.2  & 32.0  & 63.3  & 71.6  & 27.8  & 68.2  & 42.5  & 41.7  & 52.1  & 63.0  \\
    DNE   & 86.6  & 26.5  & 69.6  & 69.0  & 23.4  & 73.1  & 40.2  & 44.2  & 49.3  & 65.8  \\
    InfoBERT & 92.2  & 41.7  & 54.8  & 74.9  & 45.2  & 51.1  & 45.8  & 45.4  & 50.8  & 65.6  \\
    TAVAT & 92.2  & 40.4  & 56.3  & 74.3  & 42.3  & 54.2  & 45.7  & 42.7  & 53.8  & 64.2  \\
    FreeLB & 93.1  & 42.7  & 53.7  & 75.9  & 48.2  & 47.7  & 45.7  & 46.7  & 49.3  & 67.5  \\
    \midrule
    FreeLB++$10$ & 93.3  & 41.9  & 54.8  & 75.8  & 46.1  & 50.3  & 45.9  & 44.2  & 52.4  & 65.3  \\
    FreeLB++$30$ & \bf93.4  & 45.6  & 50.6  & 78.1  & 47.4  & 48.8  & 45.7  & 42.9  & 53.6  & 66.0  \\
    FreeLB++$50$ & 92.0  & 45.5  & 50.4  & 77.2  & 47.4  & 48.4  & 45.3  & 44.6  & 51.4  & 67.5  \\
    \midrule

    GAT$_{FGM}$ (ours) & 92.8  & 45.8  & 49.8  & 78.5  & 49.0  & 46.3  & 47.0  & 45.5  & 50.1  & 64.9  \\
    GAT$_{FreeLB++}10$ (ours) & 93.2  & 49.5  & 46.3  & 80.6  & 52.4  & 43.2  & \bf47.9 & \bf48.3  & \bf46.9  & \bf68.9   \\
    GAT$_{FreeLB++}30$ (ours) & 92.7  & \bf52.5  & \bf42.2  & \bf82.3  & \bf53.8  & \bf40.9  & 47.5   & 46.1  & 50.0  & 65.8 \\
    
    \bottomrule\bottomrule
    \end{tabular}%
      \caption{ Main defense results on the SST-2 dataset, including the test accuracy on the clean test set ({\bf Clean \%}), the robust accuracy under adversarial attacks ({\bf RA \%}), the attack success rate ({\bf ASR \%}), and the average number of queries requiring by the attacker ({\bf \# Query}).}
  \label{tab:main_re_sst}%
\end{table*}%

\subsection{Attacking Methods}
\label{subsec:atk_m}

Follow \citet{li-etal-2021-searching}, we adopt TextFooler \cite{jin-etal-2019-is-bert}, TextBugger \cite{li-etal-2018-textbugger} and BAE \cite{garg-and-ramakrishnan-2020-bae} as attackers. TextFooler and BAE are word-level attacks and TextBugger is a multi-level attacking method. We also impose restrictions on these attacks for a fair comparison, including:
\begin{itemize}
    \item[1.] The maximum percentage of perturbed words $p_{max}$
    \item[2.] The minimum semantic similarity $\varepsilon_{min}$ between the original input and the generated adversarial example
    \item[3.] The maximum size $K_{syn}$ of one word's synonym set
\end{itemize}
Since the average sentence length of IMDb and SST-2 are different, $p_{max}$ is set to 0.1 and 0.15, respectively; $\varepsilon_{min}$ is set to 0.84; and $K_{syn}$ is set to 50. All settings are referenced from previous work.

\subsection{Adversarial Training Baselines}
\label{subsec:at_baselines}

We use BERT$_{base}$ \cite{devlin-etal-2019-bert} as the base model to evaluate the impact of the following variants of adversarial training on accuracy and robustness and provide a comprehensive comparison with our proposed GAT.
\begin{itemize}[leftmargin=*]
\setlength{\itemsep}{0pt}
\setlength{\parsep}{0pt}
\setlength{\parskip}{0pt}
    \item  Adversarial Data Augmentation
    \item  ASCC \cite{dong-etal-2021-towards}
    \item  DNE \cite{zhou-etal-2021-defense}
    \item  InfoBERT \cite{wang-etal-2021-infobert}
    \item  TAVAT \cite{li-and-qiu-2021-token}
    \item  FreeLB \cite{zhu-etal-2020-freelb}
    \item  FreeLB++ \cite{li-etal-2021-searching}
\end{itemize}

ASCC and DNE adopt a convex hull during training.
InfoBERT improves robustness using mutual information.
TAVAT establishes a token-aware robust training framework. 
FreeLB++ removes the norm bounded projection and increases search steps.

We only compare GAT with adversarial training-based defense methods and leave comparisons with other defense methods (e.g., certified defenses) for future work.



\begin{table*}[htbp]\small 
\setlength{\tabcolsep}{1.2mm}
  \centering

    \begin{tabular}{lcccccccccc}
    \toprule\toprule
    \multicolumn{1}{c}{\multirow{2}{*}{\textbf{IMDb}}} & \multirow{2}{*}{\textbf{Clean \%}} & \multicolumn{3}{c}{\textbf{TextFooler}} & \multicolumn{3}{c}{\textbf{TextBugger}} & \multicolumn{3}{c}{\textbf{BAE}} \\
\cmidrule{3-11}          &       & \textbf{RA \%} & \textbf{ASR \%} & \textbf{\# Query} & \textbf{RA \%} & \textbf{ASR \%} & \textbf{\# Query} & \textbf{RA \%} & \textbf{ASR \%} & \textbf{\# Query} \\
    \midrule
    BERT$_{base}$ & 91.2  & 30.7  & 66.4  & 714.4  & 38.9  & 57.4  & 490.3  & 36.0  & 60.6  & 613.6  \\
    \midrule
    ADA   & 91.4  & 34.6  & 61.7  & 804.8  & 40.5  & 55.2  & 538.8  & 37.0  & 59.1  & 693.4  \\
    ASCC  & 86.4  & 22.2  & 73.9  & 595.9  & 27.2  & 68.0  & 415.8  & 34.7  & 59.1  & 642.2  \\
    DNE   & 86.1  & 14.9  & 82.2  & 520.2  & 17.4  & 79.3  & 336.9  & 35.4  & 57.8  & 630.4  \\
    InfoBERT & 91.9  & 33.0  & 63.9  & 694.1  & 40.4  & 55.8  & 469.9  & 37.3  & 59.2  & 619.6  \\
    TAVAT & 91.5  & 37.8  & 58.9  & 1082.6  & 48.8  & 46.9  & 695.5  & 41.2  & 55.2  & 896.7  \\
    FreeLB & 91.3  & 34.6  & 61.9  & 782.0  & 42.9  & 52.7  & 542.7  & 37.6  & 58.5  & 646.7  \\
    \midrule
    FreeLB++-$10$ & 92.1  & 39.5  & 56.8  & 817.9  & 46.4  & 49.3  & 516.5  & 41.2  & 55.0  & 682.3  \\
    FreeLB++-$30$ & 92.3  & 49.8  & 45.6  & 992.9  & 56.0  & 38.8  & 600.1  & 48.3  & 47.2  & 788.2  \\
    FreeLB++-$50$ & 92.3  & 50.2  & 45.3  & 1117.7  & 56.5  & 38.5  & 649.8  & 48.2  & 47.5  & 861.3  \\
    \midrule

    GAT$_{FGM}$ (ours)  & 91.8 & 58.3  & 36.0  & 1004.3  & 60.4  & 33.7  & 556.1  & \bf54.6  & \bf40.1  & 747.4   \\
    GAT$_{FreeLB++}10$ (ours) & 92.0  & 50.7  & 44.7  & 1093.8  & 54.7  & 40.4  & 648.9  & 50.7  & 44.7  & 908.5  \\
    GAT$_{FreeLB++}30$ (ours) & \bf92.4  & \bf59.0  & \bf35.7  & \bf1629.4  & \bf62.2  & \bf32.2  & \bf914.8  & 54.4  & 40.7  & \bf1213.6  \\

    \bottomrule\bottomrule
    \end{tabular}%
      \caption{Main defense results on the IMDb dataset.}
  \label{tab:main_re_imdb}%
\end{table*}%

\subsection{Implementation Details}
\label{subsec:imp_details}

We implement ASCC, DNE, InfoBERT, and TAVAT models based on TextDefender \cite{li-etal-2021-searching}.
We implement FGM, FreeLB, FreeLB++, and our GAT based on HuggingFace Transformers.\footnote{https://huggingface.co/transformers}
We implement ADA and FADA based on TextAttack \cite{morris-etal-2020-textattack}.\footnote{https://github.com/QData/TextAttack}
All the adversarial hyper-parameters settings are following their original papers.
All the models are trained on two GeForce RTX 2080 GPUs and eight Tesla T4 GPUs.

Regarding the training settings and hyper-parameters, the optimizer is AdamW \cite{loshchilov-and-hutter-2018-decoupled}; the learning rate is $2e^{-5}$; the number of epochs is $10$; the batch size is $64$ for SST-2 and $24$ for IMDb; the maximum sentence length kept for all the models is 40 for SST-2 and 200 for IMDb.

\subsection{Main Results}
\label{subsec:main_re}

Our proposed GAT can easily combine with other adversarial training methods. 
In our experiments, we combine GAT with FGM (GAT$_{FGM}$) and FreeLB++ (GAT$_{FreeLB++}$), respectively.
We aim to evaluate if GAT can bring improvements to the simplest (FGM) and the most effective (FreeLB++) AT methods.


We summarize the main defense results on the SST-2 dataset in Table~\ref{tab:main_re_sst}.
When GAT works with the simplest adversarial training method, FGM, the resulting robustness improvement exceeds FreeLB++$50$. 
The effectiveness and efficiency of GAT allow us to obtain strong robustness while saving many search steps.
Further combining FreeLB++ on GAT can obtain stronger robustness and outperform all other methods.

Regarding the accuracy, FreeLB++$30$ obtains the highest 93.4\%. 
GAT also significantly improves accuracy.

In addition, ADA is effective in improving robustness but hurts accuracy.
It is not surprising that ASCC and DNE suffer from significant performance losses.  
However, there is no improvement in robustness and even weaker robustness under TextFooler and TextBugger attacks than the other methods. 

\begin{table}[htbp]\small 
  \centering

    \begin{tabular}{llccc}
    \toprule\toprule
    \bf{AWS} & \bf{AT method}    & \bf{Clean \%} & \bf{RA \%} & \bf{\#Query}\\
    \midrule
    None  & None  & 92.4  & 38.5 & 44.3\\
    None  & FGM   & 92.5  & 39.6 & 44.7\\
    None  & FreeLB++30 & 93.4  & 47.4 & 45.7 \\
    \midrule
    ADA   & None  & 92.2  & 42.0 & 47.0\\
    ADA   & FGM   & 91.3  & 42.7 & 46.6\\
    ADA   & FreeLB++30 & 90.9  & 51.5 & 47.5 \\
    \midrule
    FADA  & None  & 92.7  & 44.4 & 45.8\\
    FADA  & FGM   & 92.8  & 49.0 & 47.0\\
    FADA  & FreeLB++30 & 92.7  & 53.8 & 47.5 \\
    \bottomrule\bottomrule
    \end{tabular}%
  \caption{Ablation studies on the SST-2 dataset. The attacking method is TextBugger. We only report {\bf RA~\%} and {\bf \#Query} due to the space limit. ``AWS'' means adversarial word substitution methods.}
  \label{tab:ablations}%
\end{table}%

Table~\ref{tab:main_re_imdb} shows the defense results on the IMDb dataset.
The defense performances are generally consistent with that on the SST-2 dataset.
It is worth noting that GAT$_{FGM}$ achieved an extremely high {\bf RA~\%} with a medium {\bf\#Query}, which needs further exploration.

\begin{figure*}
    \centering
    \subfigure[]{\label{fig:robust_n_steps}\includegraphics[width=0.65\columnwidth]{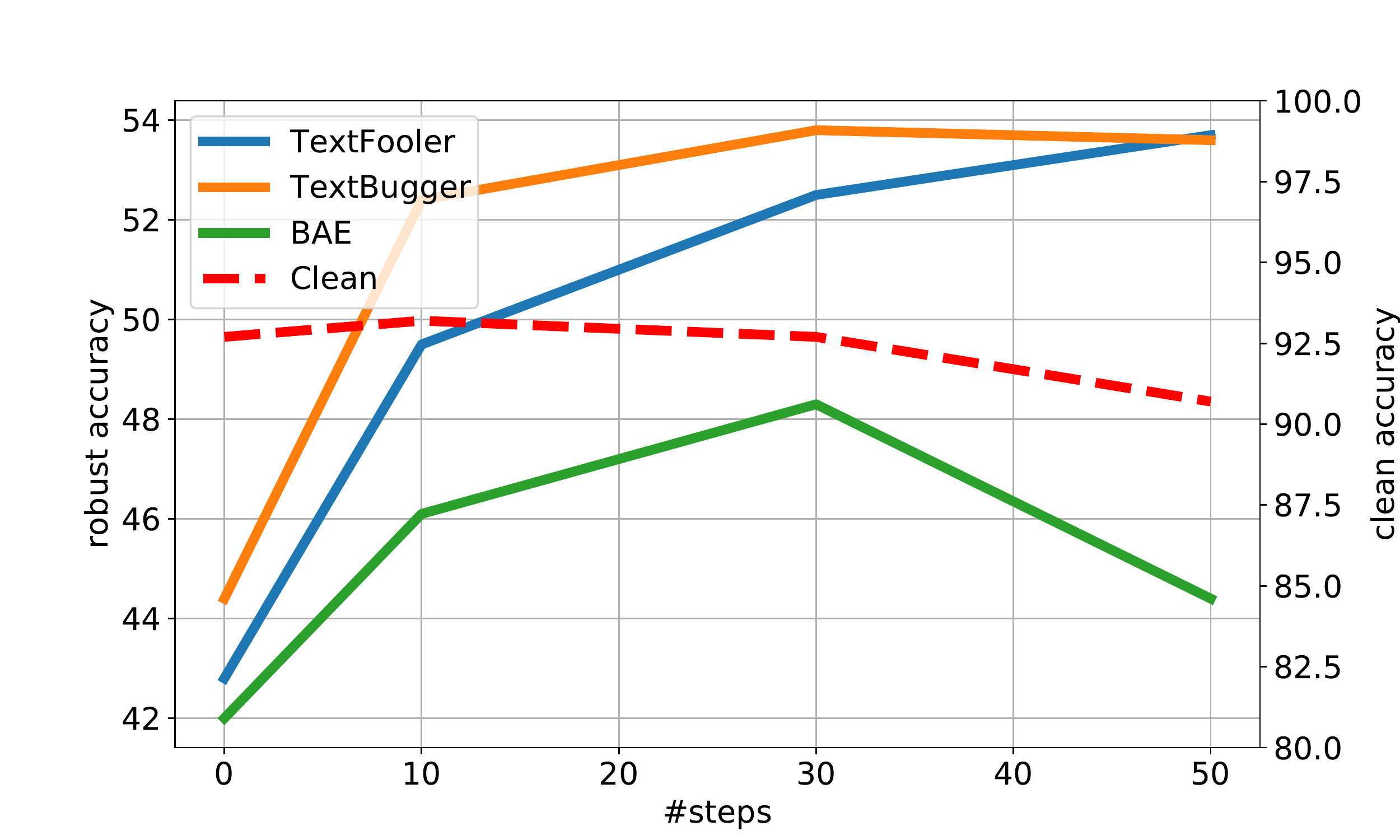}}
    \subfigure[]{\label{fig:ra_step_size}\includegraphics[width=0.65\columnwidth]{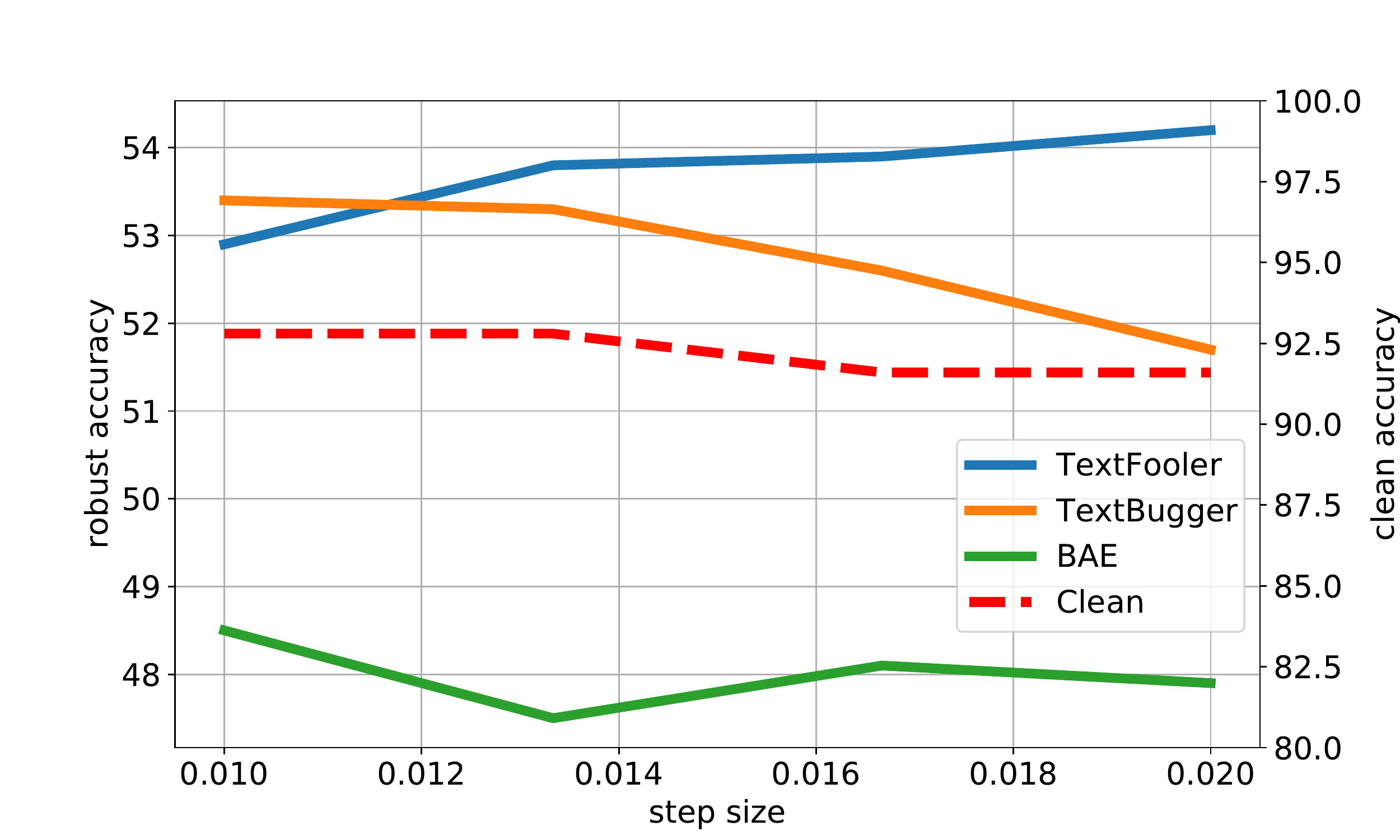}}
    \subfigure[]{\label{fig:robust_epoch}\includegraphics[width=0.65\columnwidth]{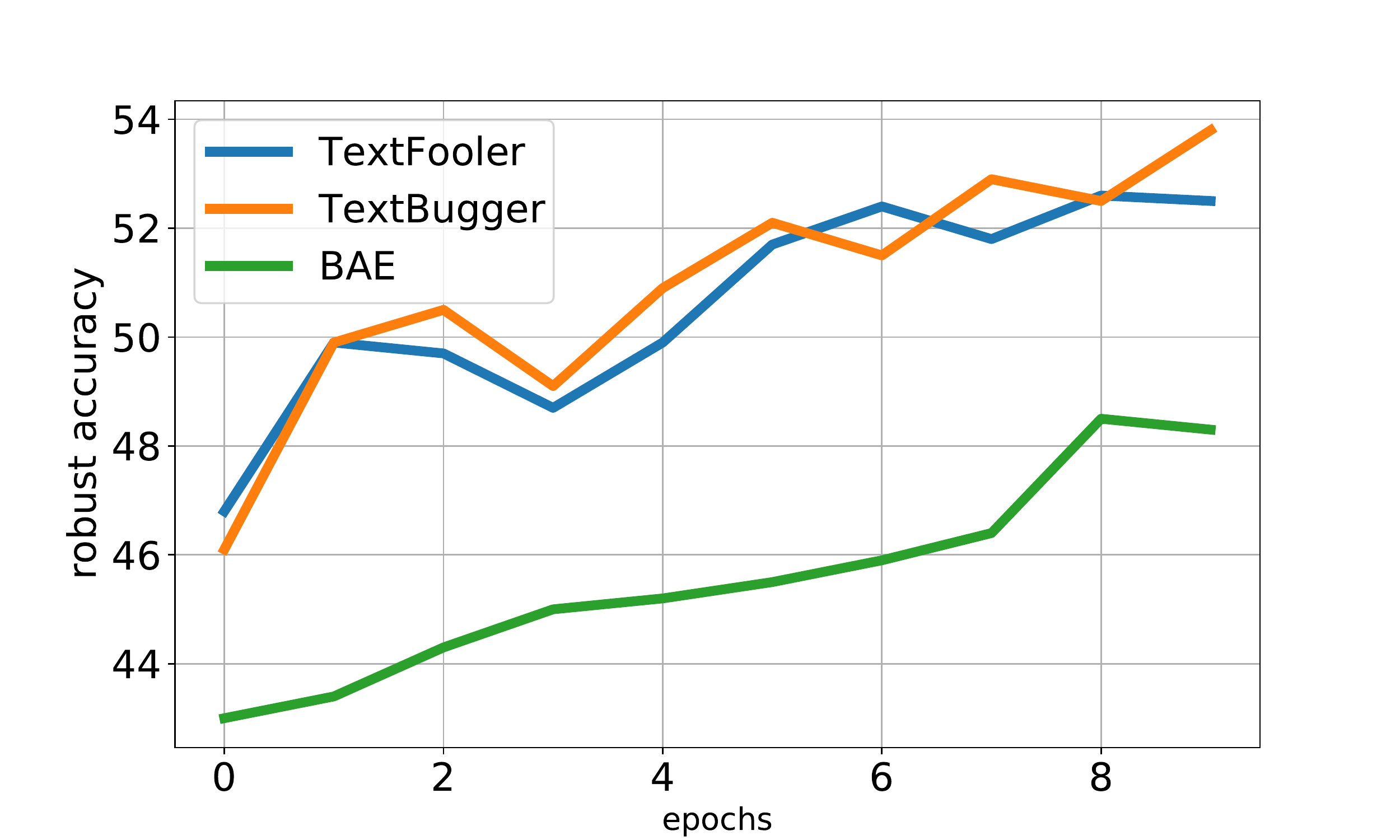}}
    \caption{(a) Robust and clean accuracy with different search steps. (b) Robust and clean accuracy with different step sizes. (c) Robust accuracy gradually increases on the SST-2 dataset during training. The adversarial training method is GAT$_{FreeLB++}30$. Zoom in for a better view.}
    \label{fig:3_fig}
\end{figure*}

\section{Discussions}
\label{sec:dis}
We further explore other factors that affect robustness and provide comprehensive empirical results.

\subsection{Ablation Studies}
\label{subsec:ablation}

We conduct ablation studies on the SST-2 dataset to assess the impact of each component of GAT.

\begin{table}[htbp]\small
\setlength{\tabcolsep}{0.4mm}
  \centering
    \begin{tabular}{lccccc}
    \toprule\toprule
    \multicolumn{1}{c}{\multirow{2}{*}{\textbf{SST-2}}} & \multicolumn{1}{c}{\multirow{2}{*}{\textbf{clean \%}}} & \multicolumn{2}{c}{\textbf{PSO}} & \multicolumn{2}{c}{\textbf{FastGA}} \\
\cmidrule{3-6}          &       & \textbf{RA \%}  & \textbf{\#Query} & \textbf{RA \%}  & \textbf{\#Query} \\
    \midrule
    BERT$_{base}$  & 92.4  & 23.9    & 322.0  & 39.2    & 234.4  \\
    \midrule
    ADA   & 92.2  & 31.4    & 348.6  & 43.2    & 268.4  \\ 
    ASCC  & 87.2  & 29.2    & 359.4  & 40.5    & 233.2 \\
    DNE   & 86.6  & 17.3    & 266.2 & 43.9     & 250.1 \\
    InfoBERT & 92.2  & 29.0     & 335.7  & 45.3    & 256.0 \\
    TAVAT & 92.2  & 25.7        & 316.2 & 42.0    & 258.7 \\
    FreeLB & 93.1  & 27.8      & 325.6   & 42.9       & 267.9                                     \\
    FreeLB++$50$ & 92.0  & 38.4  & \bf368.6  & 49.2  & 258.9  \\
    \midrule

    GAT$_{FGM}$ &92.8 &29.9 &341.0 &46.7 &275.1 \\
    GAT$_{FreeLB++}10$ &\bf93.2 &34.5 &351.3 &51.0 &289.5\\
    GAT$_{FreeLB++}30$ & 92.8   & \bf 39.7   & 359.2  & \bf 53.7    & \bf 323.9  \\
    \bottomrule\bottomrule

    \end{tabular}%
    \caption{The defense results of different AT methods against two combinatorial optimization attacks. We remove {\bf ASR~\%} due to the space limit.}
  \label{tab:sst_pso_fastga}%
\end{table}%

As shown in Table~\ref{tab:ablations}, ``FADA'' consistently outperforms ``ADA'' and ``None'' with different adversarial training methods.
Furthermore, ``FADA\&FGM'' achieve a higher {\bf RA\%} than ``None\&FreeLB++$30$'', which implies that ``FADA'' can obtain strong robustness in one adversarial search step.
``ADA'' also helps improve robustness.
However, as the number of search steps increases, so does the hurt it does to {\bf Clean~\%}.
On the contrary, ``FADA'' does not harm {\bf Clean~\%} but improves it, implying its friendliness.

\subsection{Results with Other Attacks }
\label{subsec:re_other_attacks}

We have shown that GAT brings significant improvement in robustness against three greedy-based attacks.
We investigate whether GAT is effective under combinatorial optimization attacks, such as PSO \cite{zang-etal-2020-word} and FastGA \cite{jia-etal-2019-certified}.

We can see from Table~\ref{tab:sst_pso_fastga} that GAT$_{FreeLB++}30$ obtain the highest {\bf RA~\%} against the two attacks and GAT$_{FreeLB++}10$ has the highest clean accuracy.
The results demonstrate that our proposed GAT consistently outperforms other defenses against combinatorial optimization attacks.

\subsection{Results with More Steps}
\label{subsec:re_more_steps}

As we claim in Section~\ref{sec:intro}, the accuracy should degrade with a large number of search steps.
But what happens for robustness?

We aim to see if {\bf RA~\%} can be further improved.
Figure~\ref{fig:robust_n_steps} shows that the {\bf RA~\%} gradually increases against TextFooler and TextBugger attacks.
However, {\bf RA~\%} decreases against BAE with steps more than 30, which needs more investigation.
As the steps increase, the growth rate of {\bf RA~\%} decreases, and the {\bf Clean~\%} decreases. 
We conclude that a reasonable number of steps will be good for both {\bf RA~\%} and {\bf Clean~\%}.
It is unnecessary to search for too many steps since robustness grows very slowly in the late adversarial training period while accuracy drops. 

\subsection{Impact of Step Size}
\label{subsec:impact_step_size}
A large step size (i.e., adversarial learning rate) will cause performance degradation for conventional adversarial training.
Nevertheless, what impact does it have on robustness?
We explore the impact of different step sizes on robustness and accuracy. 
As shown in Figure~\ref{fig:ra_step_size}, the clean test accuracy slightly drops as the step size increases. 
The robust accuracy under TextFooler attack increases, while the robust accuracy under Textbugger and BAE attacks decrease. 
Overall, the impact of step size on robustness needs further study.


\subsection{Impact of Training Epochs}
\label{subsec:impact_epoch}
\citet{ishida-etal-2020-zeroloss} have shown that preventing further reduction of the training loss when reaching a small value
and keeping training can help generalization.
In adversarial training, it is naturally hard to achieve zero training loss due to the insufficient capacity of the model \cite{zhang-etal-2021-geometryaware}.

\begin{table*}[ht]\small 
\setlength{\tabcolsep}{1.2mm}
  \centering
  
    \begin{tabular}{lcccccccccc}
\toprule\toprule
\multicolumn{1}{c}{\multirow{2}{*}{\textbf{SST-2}}} & \multirow{2}{*}{\textbf{Clean \%}} & \multicolumn{3}{c}{\textbf{TextFooler}} & \multicolumn{3}{c}{\textbf{TextBugger}} & \multicolumn{3}{c}{\textbf{BAE}} \\
\cmidrule{3-11}          &       & \textbf{RA \%} & \textbf{ASR \%} & \multicolumn{1}{c}{\textbf{\# Query}} & \textbf{RA \%} & \textbf{ASR \%} & \multicolumn{1}{c}{\textbf{\# Query}} & \textbf{RA \%} & \textbf{ASR \%} & \textbf{\# Query} \\
    \midrule
    RoBERTa$_{base}$ & 93.0 & 38.8 & 58.0 & 74.5 & 41.4 & 55.2 & 45.5 & 40.3 & 56.4 & 63.6 \\
    \midrule
    GAT$_{FGM}$ &91.4 &47.6 	&47.7 	&78.6 	&49.8 	&45.3 	&46.3 	&42.7 	&53.2 	&65.3 
  \\
    GAT$_{FreeLB++}30$ & 93.2   & 52.1  & 43.7  & 95.5 & 54.2  & 41.3  & 55.8  & 47.0  & 49.1  & 76.9  \\

    \bottomrule\bottomrule
    \end{tabular}%
    \caption{Defense results on RoBERTa model on the SST-2 dataset.}
  \label{tab:sst_roberta}%
\end{table*}%

\begin{table*}[ht]\small 
\setlength{\tabcolsep}{1.2mm}
  \centering
  
    \begin{tabular}{lcccccccccc}
\toprule\toprule
\multicolumn{1}{c}{\multirow{2}{*}{\textbf{SST-2}}} & \multirow{2}{*}{\textbf{Clean \%}} & \multicolumn{3}{c}{\textbf{TextFooler}} & \multicolumn{3}{c}{\textbf{TextBugger}} & \multicolumn{3}{c}{\textbf{BAE}} \\
\cmidrule{3-11}          &       & \textbf{RA \%} & \textbf{ASR \%} & \multicolumn{1}{c}{\textbf{\# Query}} & \textbf{RA \%} & \textbf{ASR \%} & \multicolumn{1}{c}{\textbf{\# Query}} & \textbf{RA \%} & \textbf{ASR \%} & \textbf{\# Query} \\
    \midrule
    DeBERTa$_{base}$ & 94.6 & 53.7 & 43.4 & 79.5 & 55.1 & 42.0 & 48.7 & 49.8 & 47.5 & 66.8 \\
    \midrule
    GAT$_{FGM}$ &94.5 	&54.6 	&42.1 	&82.6 	&57.7 	&38.8 	&50.0 	&48.9 	&48.2 	&66.7 
  \\
    GAT$_{FreeLB++}30$ &94.7 	&60.4 	&35.7 	&83.4 	&62.0 	&33.9 	&51.2 	&52.2 	&44.4 	&69.9 
  \\
    
    \bottomrule\bottomrule
    \end{tabular}%
    \caption{Defense results on DeBERTa model on the SST-2 dataset.}
  \label{tab:sst_deberta}%
\end{table*}%

Therefore, we investigate whether more training iterations result in stronger robustness in adversarial training. 
We report the {\bf RA~\%} achieved by GAT$_{FreeLB++}30$ at each epoch in Figure~\ref{fig:robust_epoch}. 
We observe that the {\bf RA~\%} tends to improve slowly, implying that more training iterations result in stronger model robustness using GAT.




\subsection{Results with Other Models }
\label{subsec:re_other_models}

We show that GAT can work on more advanced models. We choose RoBERTa$_{base}$ \cite{liu-etal-2019-roberta} and DeBERTa$_{base}$ \cite{he-etal-2021-deberta}, two improved versions of BERT, as the base models. 
As shown in Table~\ref{tab:sst_roberta} and Table~\ref{tab:sst_deberta}, GAT slightly improve robustness of RoBERTa and DeBERTa models.

\subsection{Limitations}
\label{subsec:limit}
We discuss the limitations of this work as follows.
\begin{itemize}[leftmargin=*]
    \item As we clarify in Section~\ref{subsubsec:lo}, instead of dynamically generating friendly adversarial data in training, we choose to pre-generate static augmentation. 
We do this for efficiency, as dynamically generating discrete sentences in training is computationally expensive. 
Although it still significantly improves robustness in our experiments, such a tradeoff may lead to failure because the decision boundary changes continuously during training.
    \item GAT performs adversarial training on friendly adversarial data. It may help if we consider the decision boundaries when performing gradient-based adversarial training—for example, stopping early when the adversarial data crosses the decision boundary. We consider this as one of the directions for future work.
\end{itemize}

\section{Conclusion}
\label{sec:conclusion}

In this paper, we study how to improve robustness from a geometry-aware perspective. 
We first propose FADA to generate friendly adversarial data that are close to the decision boundary. 
Then we combine gradient-based adversarial training methods on FADA to save a large number of search steps, termed geometry-aware adversarial training (GAT). 
GAT can efficiently achieve state-of-the-art defense performance without hurting test accuracy. 

We conduct extensive experiments to give in-depth analysis, and we hope this work can provide helpful insights on robustness in NLP.

\section*{Acknowledgments}
The authors would like to thank the anonymous reviewers for their helpful suggestions and comments. This work is supported in part by the National Natural Science Foundation of China under Grant No. 61902082 and 61976064,  and the Guangdong Key R\&D Program of China 2019B010136003.
\bibliography{anthology,custom}
\bibliographystyle{acl_natbib}

\end{document}